\documentclass[runningheads]{llncs}

\usepackage{eccv}

\usepackage{eccvabbrv}

\usepackage{graphicx}
\usepackage{booktabs}
\usepackage{diagbox}

\usepackage[accsupp]{axessibility}  

\usepackage[pagebackref,breaklinks,colorlinks,citecolor=eccvblue]{hyperref}

\usepackage{orcidlink}

\begin{document}

\title{FastCAR: Fast Classification And Regression
Multi-Task Learning via Task Consolidation for Modelling a Continuous Property Variable of Object Classes
} 

\titlerunning{FastCAR}

\author{Anoop Kini\ \ \and
Andreas Jansche\ \ \and
Timo Bernthaler\ \ \and
Gerhard Schneider\
}

\authorrunning{A.~Kini et al.}

\institute{Aalen University, Beethovenstraße.~1, 73430 Aalen, Germany\\ 
\email{\{anoop.kini, andreas.jansche, timo.bernthaler, gerhard.schneider\}@hs-aalen.de}
}

\maketitle

\begin{abstract}
FastCAR is a novel task consolidation approach in Multi-Task Learning (MTL) for a classification and a regression task, despite task heterogeneity with only subtle correlation. It addresses object classification and continuous property variable regression, a crucial use case in science and engineering. FastCAR involves a labeling transformation approach that can be used with a single-task regression network architecture. FastCAR outperforms traditional MTL model families, parametrized in the landscape of architecture and loss weighting schemes, when learning of both tasks are collectively considered (classification accuracy of 99.54\%, regression mean absolute percentage error of 2.4\%). The experiments performed used an Advanced Steel Property dataset \url{https://github.com/fastcandr/Advanced-Steel-Property-Dataset} contributed by us. The dataset comprises 4536 images of 224x224 pixels, annotated with object classes and hardness properties that take continuous values. With our designed approach, FastCAR achieves reduced latency and time efficiency. 

  \keywords{Multi-Task Learning \and Classification and Regression \and Label Transformation}
\end{abstract}

\section{Introduction}
\label{sec:intro}

Multi-Task Learning (MTL) networks are widely used to optimize multiple objective functions \cite{sener2018multi, di2023multi} simultaneously, using just one model. Apart from achieving reduced latency benefits \cite{guo2020learning}, MTL can prospectively learn more robust or universal representations \cite{zhang2021survey} than those with single-task models. In this way, MTL supports better generalization across all tasks via shared feature representation  \cite{zhang2021survey, liu2019end} while striving for low latency. MTL networks are widely used in computer vision for object detection and scene understanding, among many other applications involving classification and regression.

For achieving performance with MTL networks, task-shared (generalizable) feature representations and task-specific feature representations are both critical \cite{liu2019end}. The two requirements broadly translate to two experimental knobs. First, a network architecture (``how to share'') \cite{liu2019end}, which permits task-shared (generalizable) features while providing the ability to learn task-specific features. Second, but equally critical, is the multi-task loss function (``how to balance tasks?'')\cite{liu2019end}; it must enable learning all tasks without favoring only a few tasks towards learning completion while restricting others. The loss functions frequently demand weight assignment to tasks in an automated manner \cite{liu2019end, chen2018gradnorm}; the tasks can comprise both classification and regression \cite{misra2016cross, liu2019end}. In brevity, a suitable combination of network architectures and weighting schemes for loss function governs the performance of MTL networks \cite{liu2019end, misra2016cross}.

The network architecture evolution included several milestones. Hard Parameter Sharing (HPS) network architecture \cite{caruana1993multitask} involves sharing hidden layers across tasks while maintaining task-specific output layers. Cross-stitch network architecture \cite{misra2016cross} allows constituent cross-stitch units to model shared feature representation with linear combinations of feature maps that are end-to-end learnable. Multi-Task Attention Network (MTAN) architecture \cite{liu2019end} automatically learns task-shared and task-specific features using a global feature pool with task-specific attention modules. For classification and regression, these networks perform multi-task learning if the tasks are mutually related \cite{liu2019end, misra2016cross}. 

The architecture evolution has progressed further for increased immunity towards task dissimilarity. Multi-gate Mixture of Experts (MMoE) architecture learns to model task relationships \cite{ma2018modeling}. Learning To Branch (LTB) architecture \cite{guo2020learning} learns ``where to share across tasks'' or ``where to perform task-specific branching'' in the network. 

The second critical experimental knob in the MTL landscape is the weighting scheme. Each scheme, upon unveiling, considers feedback of different types.  Gradient-related feedback could be one type of weighting scheme like GradNorm \cite{chen2018gradnorm}, MGDA \cite{sener2018multi}, GradDrop \cite{chen2020just}, PCGrad (or Gradient surgery) \cite{yu2020gradient}, GradVac (or Gradient Vaccine)\cite{wang2020gradient}, CAGard (or Conflict Aversive Gradient) \cite{liu2021conflict}, MoCo \cite{fernando2022mitigating}, Aligned MTL \cite{senushkin2023independent}. Gradient feedback parameters can consist of positive curvature \cite{yu2020gradient}, average gradient \cite{yu2020gradient}, conflicting gradients\cite{yu2020gradient, chen2020just}, gradient differences\cite{yu2020gradient}, gradient magnitude similarity with cosine distance \cite{senushkin2023independent}, and gradient-based training rate adjustments for tasks \cite{chen2018gradnorm}. 

Other weighting schemes developed have other feedback mechanisms like statistical Uncertainty Weighting (UW) \cite{kendall2018multi} using homoscedastic uncertainty or Geometric Loss Strategy (GLS) \cite{chennupati2019multinet++} that is scale invariant. Some weights consider the time aspect for weighting, like the Dynamic Weight Average (DWA) \cite{liu2019end}, or even consider game theory dynamics using Nash weighting (Nash MTL) \cite{navon2022multi}.  

It is critical that the constituting individual tasks do not hurt the performance of one another, referred often as negative transfer of knowledge  \cite{guo2020learning, tang2020progressive, standley2020tasks}. The resolution frequently demands domain knowledge or manual interventions, which need not trivially resolve the complexity, given the loss function complexity and its dynamics over time. This retains the impetus for further development of architectures and weighting schemes for handling dissimilar tasks in MTL. Recently, new MTL architectures for online handwriting recognition were reported \cite{ott2022joint}, involving a classification and a regression. More specifically, a multivariate time series classification and trajectory regression. The novel architectures served to combine two heterogeneous tasks that were subtly correlated, involving classification and regression.   

We present a task consolidation approach for heterogeneous but subtly correlated tasks, specifically for object classification and regression-based property variable modeling, called FastCAR (Fast Classification And Regression). FastCAR adopts a novel label transformation strategy such that it is usable with merely single-task models. Note that this use case performs object identification but requires no localization via regression (object occupies entire frame), but rather requires regression for property modeling, which is of paramount relevance across a wide range of disciplines in scientific research and engineering. On a newly contributed Dataset called Advanced Steel Property, FastCAR handles task heterogeneity while also being computationally efficient for training and inference. The simplicity of the architecture, along with the simplicity of the labeling transformation approach, collectively achieve reduced latency in the domain of MTL. 

Our contribution is three-fold:
\begin{itemize}
\item We investigate and contribute towards task consolidation of heterogeneous tasks in multi-task deep learning, involving classification and regression, via label transformation.

\item We identify a use-case on object classification and its property regression, which obeys the above condition. The identification is expected to be very significant for a wide range of science and engineering disciplines due to the generic nature of problem formulation to involve objects occupying the entire frame and their property. 

\item We also contribute a dataset satisfying the above two criteria from the domain of properties of advanced steels with 4536 images by performing optical microscopy.
\end{itemize}

\section{Related Work}
\label{sec:RelatedWork}

\subsection{Task Consolidation of Heterogeneous Tasks with Architectures and Weighting Schemes}
\label{subsec:LiteratureTaskconsolidation}
Task consolidation between heterogeneous tasks, with no more than a subtle correlation, can be particularly challenging in multi-task deep learning \cite{guo2020learning, tang2020progressive}. Previous work for task consolidation explores possibilities in a landscape parametrized by architectures and weighting schemes for loss functions. Architectures like LTE \cite{guo2020learning} and PLE \cite{tang2020progressive} parametrized with automated weighting schemes, could still offer hopes to address this challenge, towards tolerating low task similarity. 

\subsection{Label Transformation}
\label{subsec:LiteratureLabelTransformation}

Label transformation has been utilized as another
dimension (experimental knob) for searching task consolidation possibilities involving classification and regression. Optimal Transport Assignment (OTA) \cite{ge2021ota}, Task-aligned One-stage Object Detection (TOOD) \cite{feng2021tood}, and generalized Focal Loss (GFL) \cite{li2020generalized} are some examples. However, such labeling approaches are confined to object identification with localization, which are related tasks \cite{liu2019end, misra2016cross}, and do not typically suffer from task heterogeneity. There are no evident approaches to date on labeling strategies for dealing with heterogeneous tasks involving classification and regression.

\subsection{Dataset for Property Modelling}
\label{subsec:LiteraturePropertyModelling} 
No publicly available datasets to date involve object classification and its class-specific property modeling, particularly when tasks are heterogeneous. We contribute an "Advanced Steel Property dataset" \url{https://github.com/fastcandr/Advanced-Steel-Property-Dataset} in this direction. 

Some relevant fields that property modeling of heterogeneous tasks can contribute include research disciplines in biological and chemical sciences, along with mechanical, chemical, and materials engineering, among many others.

\section{FastCAR: Fast Classification And Regression}
\label{sec:Methods}

The FastCAR labeling approach is a hybrid labeling approach, designed such that a hybrid label collectively describes the class label and regression label of each datapoint (class label of an object that occupies the entire frame), in the dataset. Both labels are traditionally essential for comparing the predictions during the progress of MTL model training.

By definition, `property’ refers to a rule that applies to all elements of a set. For a class-specific subset of the data, the regression values can vary in a class-specific range, conforming to the object property of that class. We utilized this mathematical constraint to guide the label transformation. The hybrid labels, after transformation reside in a continuous variable space.

The following 2 guidelines comprise the FastCAR label transformation algorithm:
\begin{enumerate}
\item Label transformation step: Transform the regression labels by imposing a mathematical constraint, such that the intersection of class-specific regression label range, between any two classes, is empty. As a part of the transformation, add a class-specific constant. Use the global maxima and minima of the transformed labels for centering about zero. Note the class-specific constants that were added, necessary in the event of an inverse transform.  

\item One epoch test: Use hybrid labels obtained by label transformation to train a single task regressor. The average magnitude of all gradients computed after the first epoch must be less than a constant value of 2. Otherwise, modify the class-specific constants in Step 1 until the gradients are acceptable.
\end{enumerate}

The FastCAR algorithm for label transformation allows a hybrid label to contain its respective class label information and a respective regression label information, while also incorporating gradient feedback from the network. This label information compression approach can be used with a simple architecture for training, particularly a regression network, without demanding a task-specific decoder branch. With inverse transform, which involves subtracting class-specific constants from hybrid labels, enables classification and regression labels to be back-calculated.   

\begin{figure}[tb]
  \centering
  \begin{subfigure}{0.4\linewidth}
  \centering
   \includegraphics{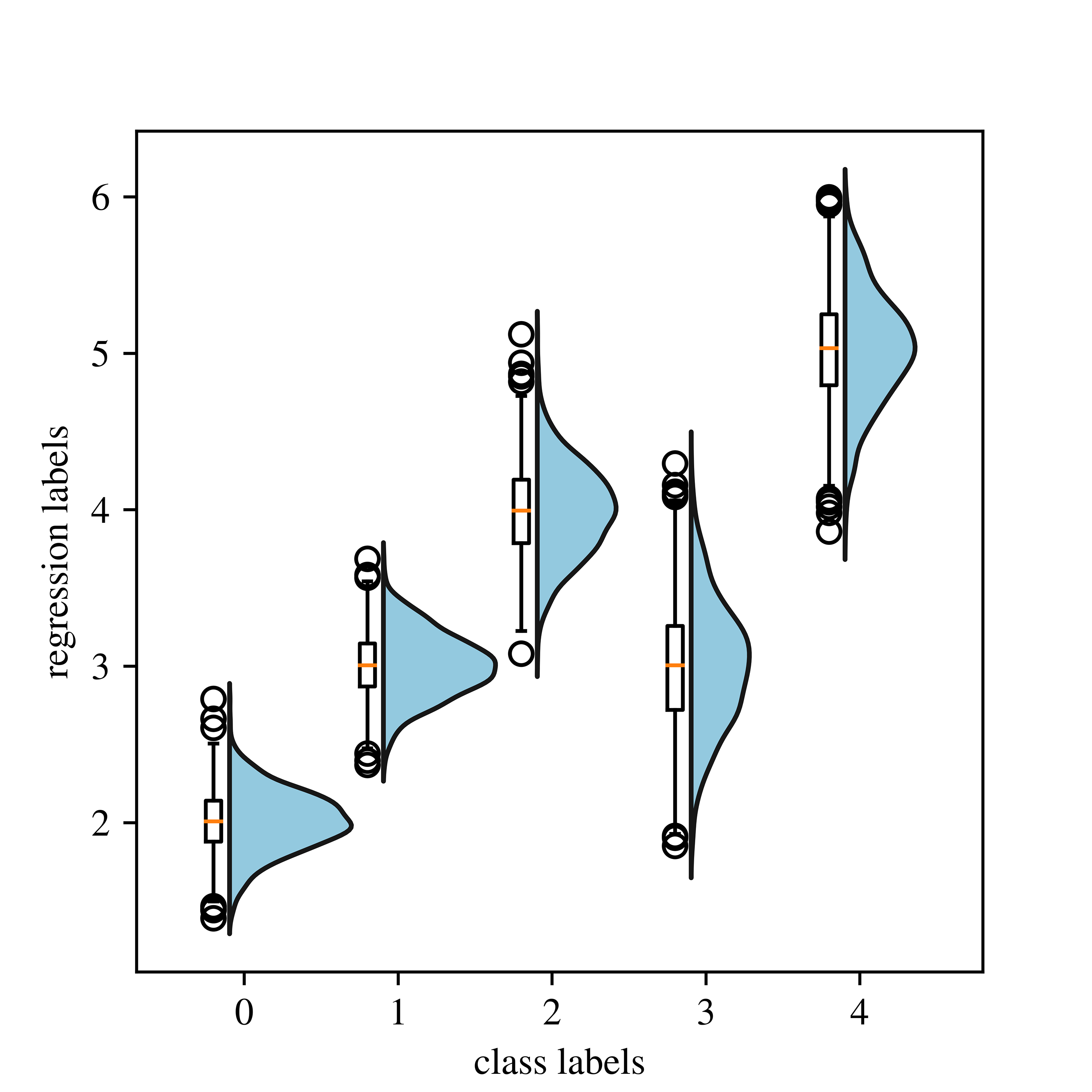}
    \caption{The regression labels are constrained for an object class, by definition of an object property}
    \label{fig:short-a}
  \end{subfigure}
  \hfill
  \begin{subfigure}{0.45\linewidth}
  \centering
   \includegraphics{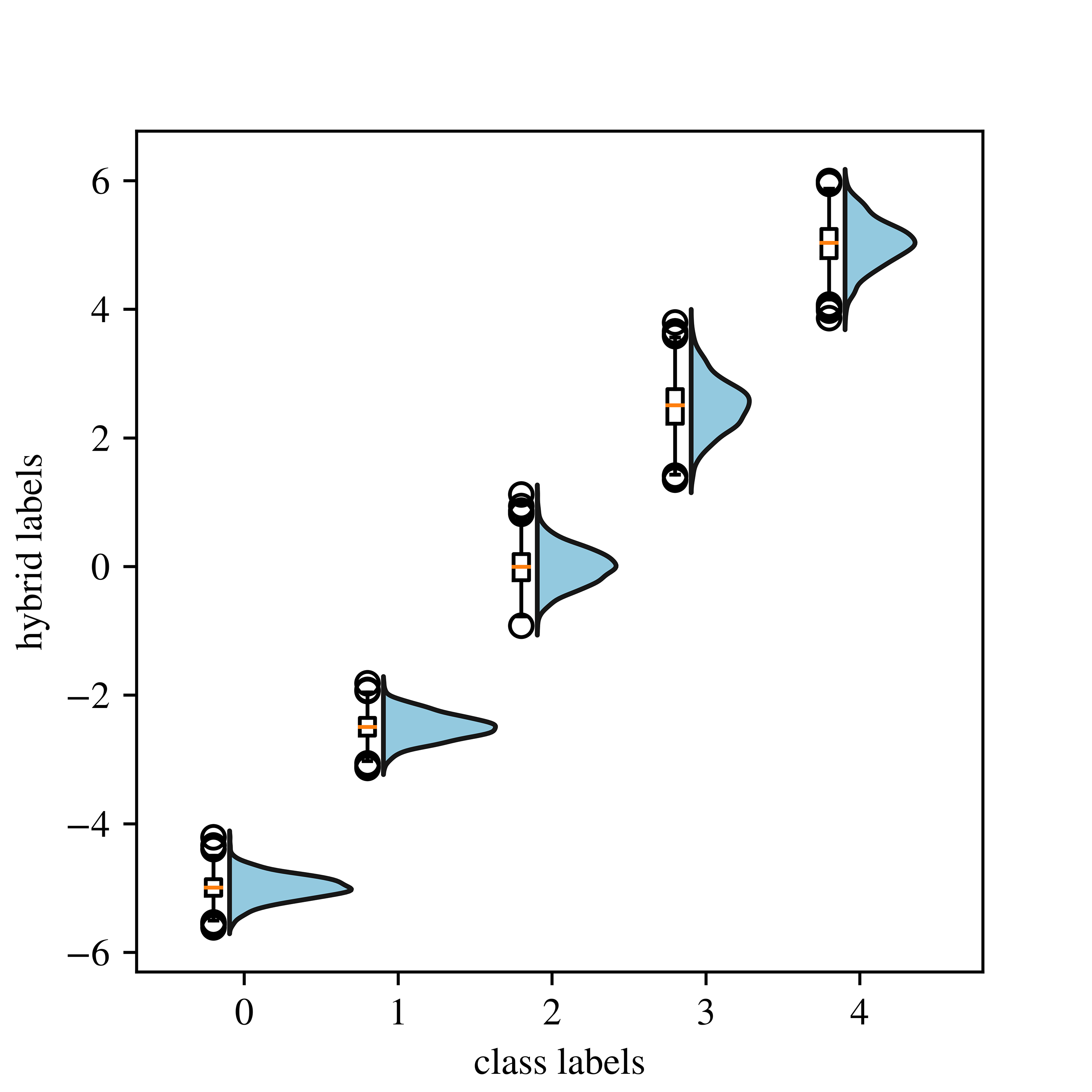}
    \caption{A hybrid label, obtained upon transformation, contains the information of the classification label and the regression label.}
    \label{fig:short-b}
  \end{subfigure}
  \caption{Schematic representation of labels in 
 a class-wise manner using a box plot and its distribution represented using a violin plot. The x-axis indicates class label indices. The regression labels and transformed labels (hybrid labels), both residing in a continuous but distinct variable space. Hybrid labels are used in conjunction with single-task regression networks as a part of the FastCAR approach for task consolidation while performing a classification and a regression task.}
  \label{fig:short}
\end{figure}

\section{Experiments}
\label{sec:blind}
\subsection{Experimental Settings}
\subsubsection{Dataset.}
We created a dataset, relevant to the property modeling of each object class, where the property takes continuous values from an advanced steel domain, called the Advanced Steel Property dataset, available with the link \url{https://github.com/fastcandr/Advanced-Steel-Property-Dataset}. The modeling involves a classification task to identify the object class and a regression task to model a continuous property variable without any localization while performing the two tasks. Each image was annotated with a classification label and a regression label. More specifically, the object classes indicate different steel types, particularly their microstructures (occupying the entire image frame), while the continuous property variable is hardness in the present work for the problem statement. 

The image acquisition scale and relevant scale for the measurement of object property must match; it was the microscopic length scale in this case. This ensured that any object localization was redundant. The images for the dataset were acquired using an optical microscope. The dataset contained 4536 acquired images (resolution 224x224 pixels), a dataset size that belongs to the typical size range in the domain of advanced steels \cite{lupulescu2015asm} of about 4000. A dataset that satisfied the criteria of size, annotated with classification and regression labels, for heterogeneous tasks without further localization requirement, and the model property was not publicly available previously. 

The dataset contained six object classes while maintaining class balance (an equal number of images for each class). Post-image acquisition, the property measurement using destructive materials testing via hardness measurements was conducted. These measurements served as the regression labels. An allied benefit of using trained networks for property prediction is the possibility of non-destructive and non-invasive property evaluation, plausibly relevant across science and engineering disciplines. The link to the dataset: available after review. 

The dataset was split into training, validation, and testing sets in the ratio 5:1:1 while maintaining class balance. The models were trained on an NVIDIA RTX\textsuperscript{TM} A6000 workstation with a dedicated GPU memory of 48 GB. 

\subsubsection{Evaluation Metrics.}

For evaluating FastCAR and the benchmark models, classification accuracy and the mean squared error (MSE error) were used to assess the learning progress of both tasks. To meet this purpose, all models were granted 100 epochs, sufficient for reaching convergence. While striving to achieve this, total execution time was noted, as a measure of time efficiency. 

The benchmark MTL experiments consisted of multiplicative combinations of 6 encoder architectures and 17 weighting schemes for the loss functions, resulting in 102 multi-task learning (MTL) models per family. The 6 architectures of encoder comprise HPS \cite{caruana1993multitask}, Cross stitch \cite{misra2016cross}, MMoE \cite{ma2018modeling}, MTAN \cite{liu2019end}, CGC \cite{tang2020progressive}, and LTB \cite{guo2020learning}. 

A total of 17 weighting schemes for parametrization include Equal Weighting (EW), Gradient Normalization (GradNorm) \cite{chen2018gradnorm}, Multiple Gradient Descent Algorithm (MGDA) \cite{sener2018multi}, Uncertainty Weights (UW) \cite{kendall2018multi}, Dynamic Weight Average (DWA) \cite{liu2019end}, Geometric Loss Strategy (GLS) \cite{chennupati2019multinet++}, Gradient sign Dropout (GradDrop) \cite{chen2020just}, Projecting Conflicting Gradients (PCGrad) \cite{yu2020gradient}, Gradient Vaccine (GradVac) \cite{wang2020gradient}, Impartial Multi-Task Learning (IMTL) \cite{liu2021towards}, Conflict-Averse Gradient descent (CAGrad) \cite{liu2021conflict}, Nash MTL \cite{navon2022multi}, Random Loss Weighting (RLW) \cite{lin2021reasonable}, Multiple objective Convergence (MoCo) \cite{fernando2022mitigating}, and Aligning of orthogonal components (Aligned MTL) \cite{senushkin2023independent}. Note that the MGDA scheme included 3 variants `loss', `loss +', and `L2'. 

Two MTL model families for benchmarking correspond to the following two decoder configurations. The Decoder-1 contains one more fully connected (fc) layer with 512 neurons (along with 1D batch normalization, ReLU activation function, and dropout of 0.5), compared to a ResNet-18 pre-trained network without the backbone. The additional layer lies between the already present adaptive average pooling layer and the fully connected layer. 

The Decoder-2, compared to Decoder-1, contains two more convolutional blocks (each followed by batch normalization and ReLU) located before the adaptive average pooling layer. The convolutional blocks have feature maps of sizes 4×4 followed by 2×2, with channels doubling across each convolutional block. After the pooling layer, yet another fully connected layer was placed.   

The experimental design focussed on serving a dual purpose. First, benchmarking of the FastCAR model performance. Second, to search and identify MTL models that can perform despite possible heterogeneous behavior between tasks (which are often challenging to determine explicitly), by attempting various combinations of architectures and weighting schemes without an explicit assignment. 

\subsubsection{Implementation Details.}
\label{subsubsec:Implementation}

The FastCAR utilizes a ResNet-18-based regression network coupled with the algorithm for hybrid labeling from \cref{sec:Methods}. The entire network was allowed to participate in the training due to the possibility of not affecting the time efficiency, owing to a simpler architecture than the MTL models. MTL models for benchmarking comprised the ResNet-18 backbone for feature extraction, used by various encoder architectures. ResNet-50 and further deep ResNet networks were not attempted as backbones because of the computational expense of running sets of 102 models per family. All MTL models utilized the encoder architectures and the loss weighting schemes from the LibMTL package \cite{lin2023libmtl}. The gradients of shared parameters of the encoder agreed with the default setting in the package, which was set to zero. The decoder architectures were subjected to gradient updates for parameter updation.    

The training of FastCAR and MTL families used an Adam optimizer with a learning rate of 1e-3 and a weight decay of 1e-4. A scheduler of type `reduced learning rate on plateau' was used with a factor of 0.1, patience of 5, and verbose as True.    

\subsection{FastCAR Performance Benchmarking with that of MTL Model Family with Decoder-1}

\begin{figure}[tb]
  \centering
  \includegraphics[height=5.65cm]{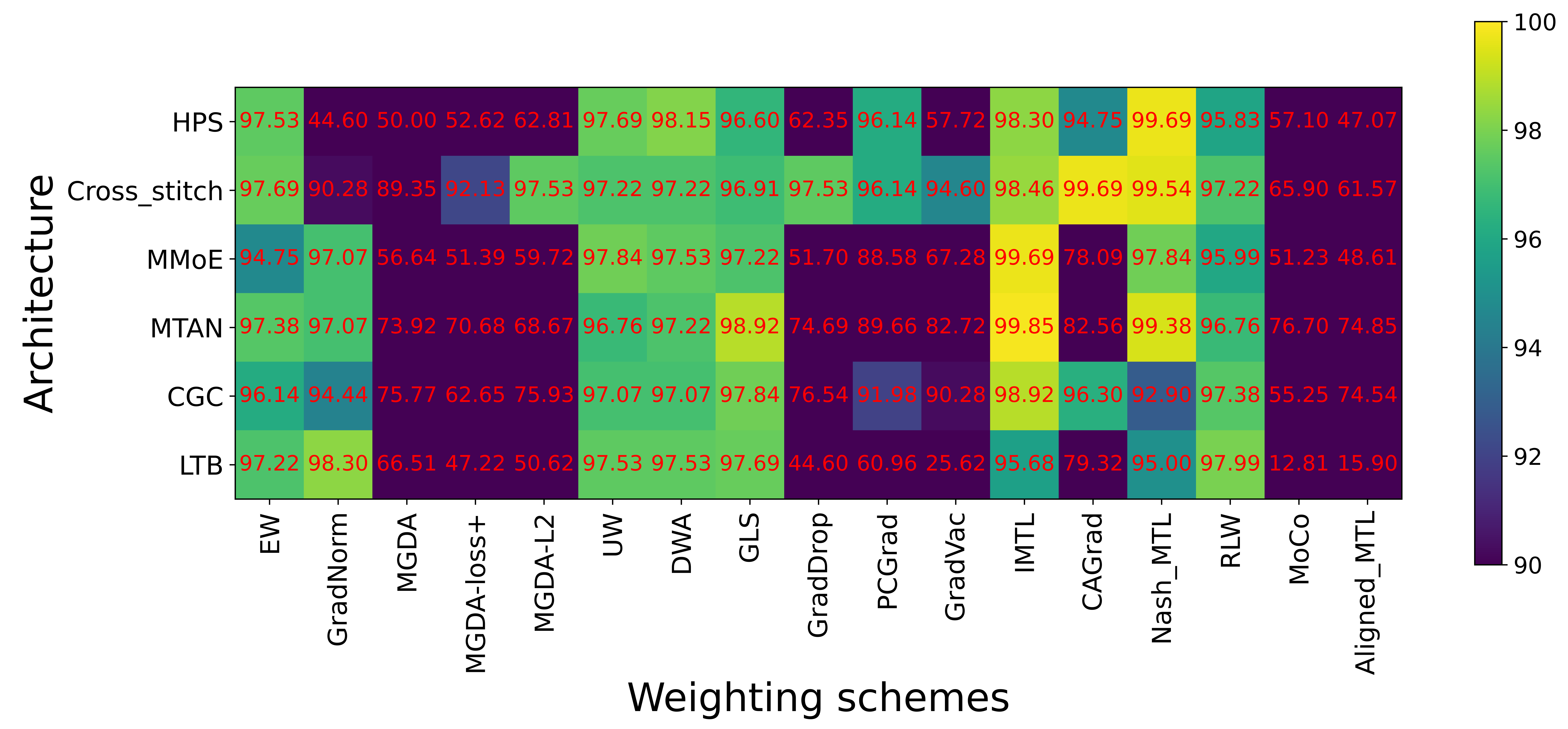}
  \caption{Performance of multi-task learning model family with Decoder-1 on the classification task (in terms of classification accuracy in \%) across the landscape of architecture \cite{misra2016cross,liu2019end, guo2020learning, ma2018modeling, caruana1993multitask, tang2020progressive} and weighting scheme \cite{liu2021conflict, liu2019end, yu2020gradient, wang2020gradient, chen2018gradnorm, senushkin2023independent, chen2020just, fernando2022mitigating, navon2022multi, sener2018multi, kendall2018multi, chennupati2019multinet++, lin2021reasonable, liu2021towards}. The FastCAR has a classification accuracy of 98.45\% which closely approaches the best model in the landscape associated with 99.85\%.}
  \label{fig:classification_v1}
\end{figure}

The \cref{fig:classification_v1} shows a 2D color contour plot (heatmap) depicting the classification accuracy of the MTL model family with Decoder-1. In the heatmap, the lighter color shades indicate higher classification accuracy than the darker color shades. 

For the classification task, there is no unanimous architecture that dominates to obtain high classification accuracy by remaining independent of the weighting scheme. The weighting schemes like EW, UW, DWA, GLS, IMTL, and RLW perform better than other schemes, across all architectures, with greater than 95\% classification accuracy on the test set. The attention mechanism-based MTAN architecture with IMTL weighting achieved the highest classification accuracy of 99.85\%. The performance by MMoE with IMTL, Cross stitch with CAgrad, and HPS with Nash MTL followed next, each with 99.69\%.

\begin{figure}[tb]
  \centering
  \includegraphics[height=5.65cm]{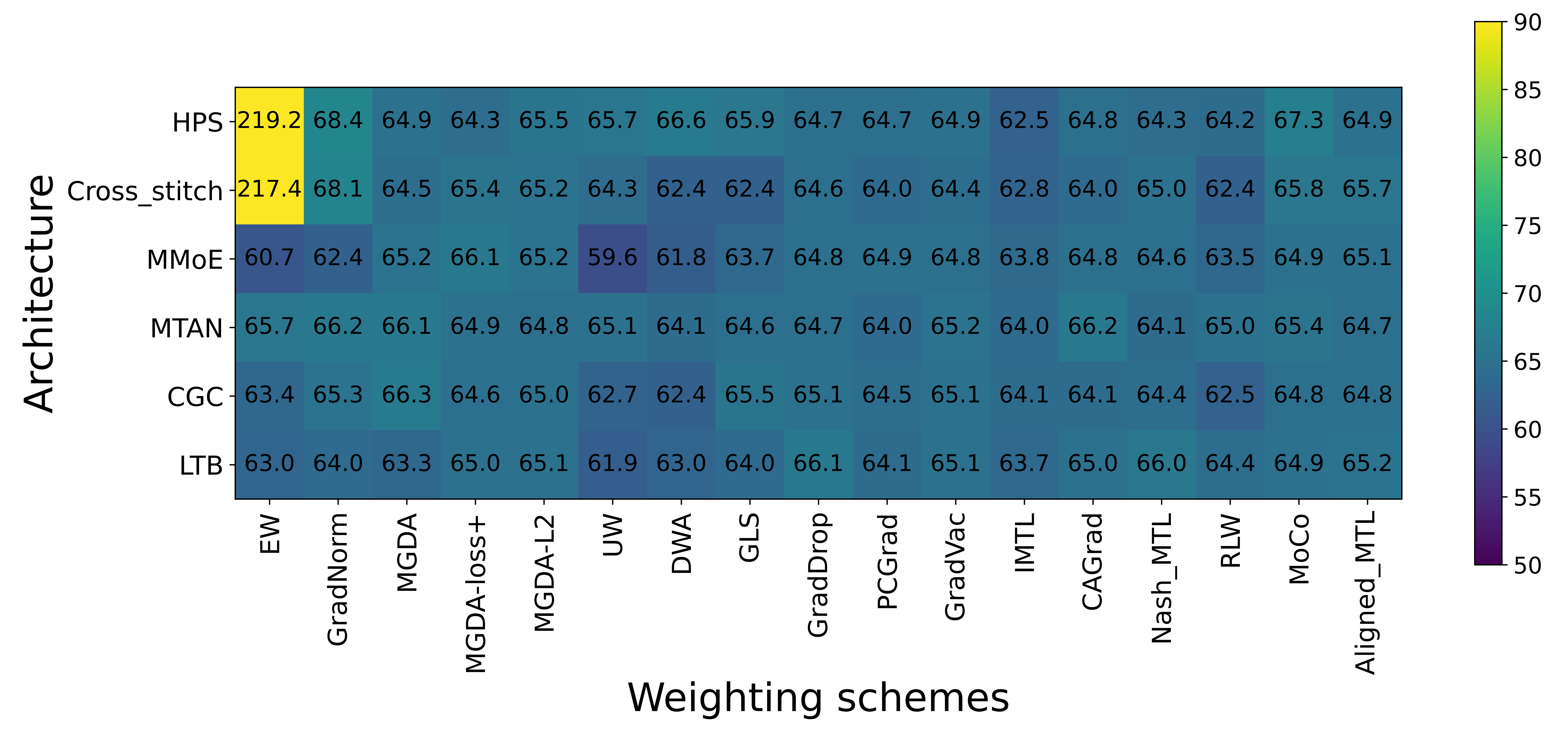}
  \caption{Performance of multi-task learning model family with Decoder-1 on the regression task (in terms of MSE error expressed as multiples of 1000) across the landscape of architecture \cite{misra2016cross,liu2019end, guo2020learning, ma2018modeling, caruana1993multitask, tang2020progressive} and weighting scheme \cite{liu2021conflict, liu2019end, yu2020gradient, wang2020gradient, chen2018gradnorm, senushkin2023independent, chen2020just, fernando2022mitigating, navon2022multi, sener2018multi, kendall2018multi, chennupati2019multinet++, lin2021reasonable, liu2021towards}. The error when converted to mean absolute percentage error is 53.7\% for the entire family, with most models approaching this value, with the few remaining models associated with even higher error. The FastCAR has a mean absolute percentage error of 2.4\%.}
  \label{fig:regression_v1}
\end{figure}

The regression loss (MSE error), expressed as multiples of 1000 in \cref{fig:regression_v1}, reveals that none of the combinations for architecture and weighting schemes enable learning on the regression task. Among the combinations, the MMoE architecture coupled with UW weighting has an MSE error of 59.6 K on the test set, the lowest error among the MTL models. Most combinations have an MSE error between 60 K and 68.4 K, except EW weighting that exceeds 200 K with HPS and Cross-stitch architectures. More specifically, regression predictions of the best MTL model vary in a narrow range from 600 to 620; note that the regression labels range from 179 to 1146. The error expressed in mean absolute percentage error is 55.76\%

For the regression task, FastCAR outperforms all MTL models with an MSE error of merely 606, which is two orders of magnitude lower than those for MTL models. When expressed in mean absolute error, it is only 12 and a mean absolute percentage error of 2.4 (98.6\% of the data is within 8\% mean absolute percentage error). The error also closely approaches the property measurement error, indicating good performance on the regression task. FastCAR also achieved a classification accuracy of 99.54\% despite an absence of explicit weighting assignment to individual tasks. Holistically, considering both tasks, FastCAR performs better than the MTL models with Decoder-1.

\subsection{FastCAR Performance Benchmarking with that of MTL Model Family with Decoder-2}

\begin{figure}[tb]
  \centering
  \includegraphics[height=5.65cm]{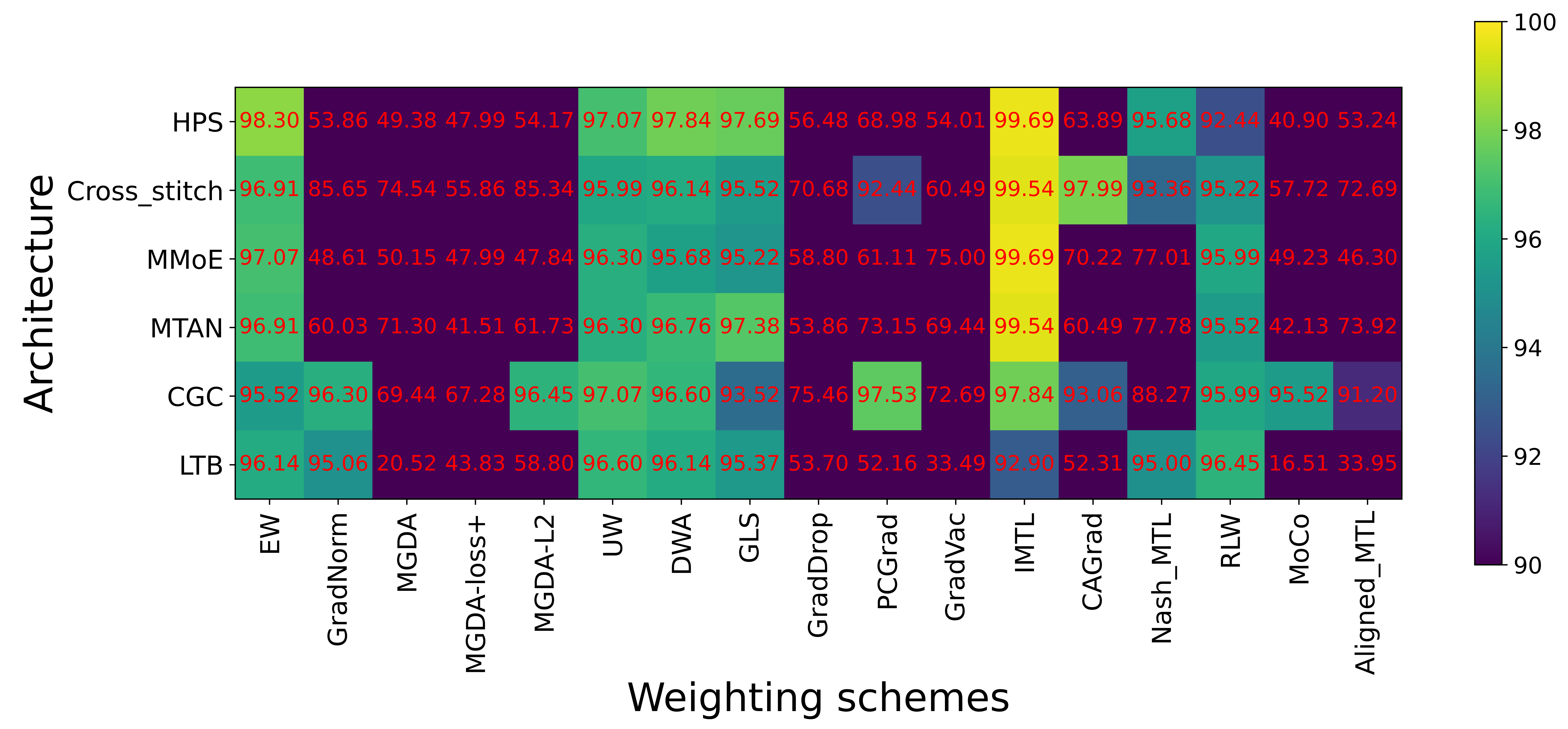}
  \caption{Performance of multi-task learning model family with Decoder-2 on the classification task (in terms of classification accuracy in \%) across the landscape of architecture \cite{misra2016cross,liu2019end, guo2020learning, ma2018modeling, caruana1993multitask, tang2020progressive} and weighting scheme \cite{liu2021conflict, liu2019end, yu2020gradient, wang2020gradient, chen2018gradnorm, senushkin2023independent, chen2020just, fernando2022mitigating, navon2022multi, sener2018multi, kendall2018multi, chennupati2019multinet++, lin2021reasonable, liu2021towards}. The FastCAR has a classification accuracy of 98.45\% which closely approaches the best model in the landscape associated with 99.69\%.}
  \label{fig:classification_v2}
\end{figure}

\begin{figure}[tb]
  \centering
  \includegraphics[height=5.65cm]{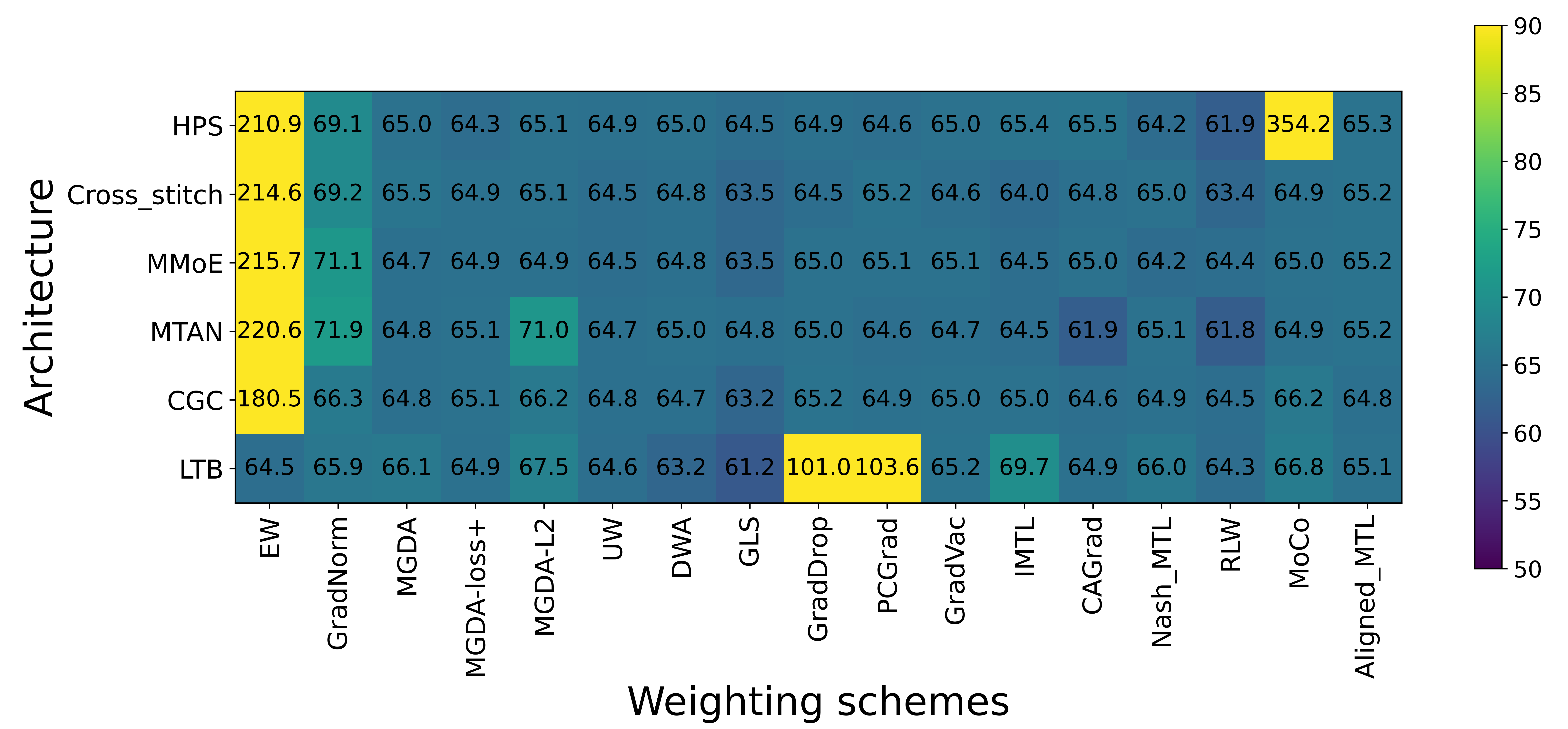}
  \caption{Performance of multi-task learning model family with Decoder-2 on the regression task (in terms of MSE error expressed as multiples of 1000) across the landscape of architecture \cite{misra2016cross,liu2019end, guo2020learning, ma2018modeling, caruana1993multitask, tang2020progressive} and weighting scheme \cite{liu2021conflict, liu2019end, yu2020gradient, wang2020gradient, chen2018gradnorm, senushkin2023independent, chen2020just, fernando2022mitigating, navon2022multi, sener2018multi, kendall2018multi, chennupati2019multinet++, lin2021reasonable, liu2021towards}. The error when converted to mean absolute percentage error is 54.2\% for the entire family, with most models approaching this value, with the few remaining models associated with even higher error. The FastCAR has a mean absolute percentage error of 2.4\%.}
  \label{fig:regression_v2}
\end{figure}

From a data compression perspective, across consecutive layers (convolutional blocks) of the ResNet-18 network, the compression factor does not exceed 2 (50\% compression); the feature maps size decreases by two times along height and width, despite an increase in channels by two times. However, after layer 4 (as designated in ResNet-18), there is substantial data compression by 49 times (~98\% compression) during adaptive average 2D pooling. The feature map size decreases to 1/7\textsuperscript{th} of height and width (feature map size decreases to 1/49\textsuperscript{th}) with an unaltered number of channels, before entering the adaptive average pooling. Adaptive average pooling often eliminates topological constraints due to mismatching dimensions of successive convolutional blocks \cite{van2019evolutionary}. It is unclear if such pooling, particularly in Decoder-1, can retain important information from the shared feature representation despite substantial compression, particularly in the context of MTL.

The Decoder2 configuration, which compared to the ResNet-18 backbone architecture, restricts the data compression factor to less than 2, even after layer 4. In other words, more than 50\% compression per layer gets prevented during the adaptive average pooling. In addition to the increased potential to prevent information loss during compression, Decoder-2 provides additional learnable parameters. 

For the classification task, as seen in \cref{fig:classification_v2} the Decoder-2, as compared to the Decoder-1, did not show a substantial difference in classification accuracy. Some combinations performed better with Decoder-2 than those with Decoder-1, while others suffered. The weighting schemes EW, UW, DWA, GLS, IMTL, and RLW, across architectures, showed a performance difference of within ±4\% classification accuracy.

The IMTL weighting with HPS and MMoE architectures achieved 99.69\%. The best classification performance is mildly lower than that of the MTL family with Decoder-1. The IMTL weighting, with Cross stitch and MTAN architectures, resulted in an accuracy of 99.54\%. 

For the regression task, Decoder-2 in \cref{fig:regression_v2}, compared to Decoder-1, showed marginal performance changes, with a relative difference of MSE loss to be within ±10 K in most cases. But, GradDrop and PCGrad weighting, particularly with the LTB architecture increased the MSE loss further by 35 and 40 K. The EW weighting coupled with either MMoE, MTAN, or CGC architectures reached an MSE loss of 180-220 K. The MoCo weighting with HPS showed a massive MSE loss, reaching 354 K.

In brevity, despite more learnable parameters in Decoder 2, the configuration in conjunction with the shared feature representation does not succeed in learning the regression task across the architecture and weighting scheme landscape. The best performance of the 2 families of the MTL models and the FastCAR are summarized in \cref{tab:Performance}

\subsection{Why FastCAR Performs Better than MTL Model Families? }
The use cases on object detection-localization \cite{ott2022joint} and object property modeling share a commonality. The classification and regression tasks can look at similar spatial regions of interest (ROI) in an image. The latter allows the ROI to extend to the entire frame. A shared feature representation cannot be directly ruled out, due to the possible commonality. 

The MTL model families, do not succeed in learning both classification and regression tasks, despite parametrizing in the landscape of architecture and weighting scheme, with 204 models. From an architectural viewpoint, even complex architectures with attention modules like MTAN cannot play a sufficiently supporting role to offer shared representations that are acceptable for weighting schemes. 
The weighing schemes also can not offer enough to the relationship with the architectures to harmoniously learn both tasks. One explanation could be the inability to learn the relative scale of the weights between classification and regression tasks. The Geometric Loss Scheme (GLS) has the potential to be scale-invariant, involving both classification and regression, which also does not help. 

The property modeling of object classes can inherently be associated with task heterogeneity. The two tasks need not honor the possibility of necessarily sharing the entire spatial ROI in the image, for feature representation \cite{standley2020tasks}. The feature space also need not be similar and can lack a shared representation. The LTB and CGC architectures, although can learn task relationships to a certain extent, it appears that the task complexity is beyond the comprehension of the architecture and weighting landscape for any meaningful shared description of features. Architectural complexity, for instance, attention mechanism-based architectures, do not necessarily imply the potential to solve the complexity of task heterogeneity. These are two different dimensions of complexity for permitting a direct comparison. Note that the MTAN architectures reported are for related tasks. 

MTL for object classification and regression property modeling FastCAR demonstrates good potential. We draw attention to an additional experimental knob of label transformation for task consolidation of heterogeneous but subtly correlated classification and regression-based tasks. The identification of use-case on property modeling of object classes carries relevance in scientific research and engineering disciplines. 

Although further optimization of the label transformation for performance improvement is possible, the focus of this work is unveiling a novel approach for task consolidation via label transformation. The former can comprise a variation in the interclass distance such that the distributions are wide apart to increase classification accuracy without affecting regression. The tradeoff, however, is that the wide range runs into the trouble of high gradients, despite the possibility of inverse transformation, which affects learning due to near-exploding grades. 

Despite neither explicitly requiring the weight assignment for the two loss functions nor requiring a robust encoder architecture, FastCAR negotiates the challenge of learning both the classification and the regression task, even if they are not similar. FastCAR holds the potential for task consolidation of a classification and a regression task, involving object property classification and property regression without localization, via label transformation and a single-task deep learning regressor network.

\begin{table}
    \centering
    \caption{Comparison of best performance among models belonging to each muti-task learning (MTL), on the classification and regression task. This has been used as a benchmark for comparison of the performance of the FastCAR. Note that the run times have also been compared.}  
    \label{tab:Performance}
  \begin{tabular}{|l|c|c|c|}
        \toprule
        \diagbox[height=2.8\line,width=6cm]{Evaluation Metric}{Model} & \shortstack{MTL \\ Family1} & \shortstack{MTL \\ Family2} & \shortstack{FastCAR }\\ 
        
        \midrule
        \multicolumn{1}{|l|}{\shortstack[l]{Classification Performance \\ (Accuracy [\%])}} & 99.85 & 99.69 & 99.54 \\    
        \midrule
          
        \multicolumn{1}{|l|}{\shortstack[l]{Regression Performance \\ (Mean Absolute Percentage Error, MAPE)}} & 53.7 & 54.2 & 2.4 \\
        \midrule
        

  \shortstack{Run time  \\ (Minutes)} &40.39±23.71  & 43.10±23.92 & 16.32  \\
          \bottomrule
    \end{tabular}

\end{table}

\subsection{How Fast is FastCAR?}

For benchmark MTL model families, the execution time for each model was considered for describing the mean and standard deviation of that family. 

The FastCAR records 16.32 min. The MTL model family with Decoder-1 takes 40.39±23.71 min, while the family with Decoder-2 needs 43.10±23.93 min. The time taken by FastCAR closely approaches the fastest MTL models that run in 15.89 min; the time corresponds to HPS architecture with either UW weighing or DWA weighting from family-1. All models with Decoder-2, compared to Decoder-1, require more time; the models run 2.7 ±1.5 min slower due to the two additional convolutional blocks in the Decoder-2.  

Note that only FastCAR succeeds in learning the regression task also, in addition to the classification task, portraying its collective performance and speed ahead of MTL models.

\subsection{Why is FastCAR Fast?}
The combined design choice of the labeling strategy and use of a single-task deep learning model contribute to the speed of the FastCAR. Let us unpack the FastCAR labeling strategy that involves 4 phases. 
\begin{enumerate}

\item Initial label definition.

\item Gradient feedback by running a single task model with such labels for one epoch. Decision-making based on the gradient feedback for label modification until favorable gradient feedback (i.e.\ ensuring the prevention of gradient explosion). 

\item Training the single task model for remaining epochs. 

\item Reverse mapping of predictions from the hybrid label space back to regression and classification labels. 
\end{enumerate}
The 1\textsuperscript{st} and 4\textsuperscript{th} steps are computationally inexpensive, involving defining and running linear equations in a back-and-forth manner during training and inference. The 2\textsuperscript{nd} step requires only one epoch for label modification per iteration to ensure that the average of the gradients for the first epoch is less than a threshold value of 2, determined by sensitivity analysis. Such label iterations are of much lower complexity than automatic or even manual weight assignment to loss functions in MTL, which frequently demands a greater number of epochs. Furthermore, the 3\textsuperscript{rd} step is also computationally fast on account of the simplicity of the network architecture of the single-task model concerning MTL architectures. The latter can comprise complex encoder architectures, while not ignoring the decoder architecture complexity, thus making it computationally expensive.  

\section{Conclusions}

FastCAR performs task consolidation for a classification and a regression task in multi-task learning, despite task heterogeneity with only subtle correlation. The identified use case applies to object classification and its continuous property modeling. Such a use case is relevant for a wide range of scientific and engineering disciplines, where property modeling is highly significant. For this, we utilized a mathematical constraint using the property definition of object classes. 

For label transformation, we considered an invertible labeling strategy by ensuring that the intersection between any two class-specific regression label ranges is null, in the hybrid label space. The labels additionally consider gradient feedback, a feedback that was frequently used for weighting scheme evolution in the architecture-weighting scheme landscape. 

FastCAR learns the object classification task (classification accuracy of 99.45\%) and property modeling regression task (mean absolute percentage error of 2.4\%), demonstrating the possibility of learning on both dissimilar tasks. Note that the MTL model families, each parameterized in the landscape of architecture and loss weighting schemes, have not been able to learn the two dissimilar tasks. 

FastCAR is time efficient, owing to the computationally inexpensive labeling strategy and its feasibility for combining with simple architectures, specifically a deep learning regression network. The two reasons allow good prospects for reduced latency. 


%
%
\bibliographystyle{splncs04}
\bibliography{main}
\end{document}